\pdfoutput=1
\documentclass{article}

\usepackage{microtype}
\usepackage{graphicx}
\usepackage{booktabs}
\usepackage{hyperref}
\usepackage{amsmath}
\usepackage{amssymb}
\usepackage{mathtools}
\usepackage[capitalize,noabbrev]{cleveref}
\usepackage[accepted]{icml2026}


\icmltitlerunning{Task-Restricted Symmetries in Recurrent Weight Space}

\newcommand{\fvu}{\mathrm{FVU}}
\newcommand{\RR}{T_{RR}}
\newcommand{\CR}{T_{C\rightarrow R}}
\newcommand{\CC}{T_{CC}}

\begin{document}

\twocolumn[
  \icmltitle{Task-Restricted Symmetries in Recurrent Weight Space}

  \begin{icmlauthorlist}
    \icmlauthor{Simon Dr{\"a}ger}{salk}
  \end{icmlauthorlist}
  \icmlaffiliation{salk}{Salk Institute for Biological Studies, La Jolla,
  CA, USA}
  \icmlcorrespondingauthor{Simon Dr{\"a}ger}{sfdraeger@gmail.com}
  \icmlkeywords{weight-space symmetries, recurrent neural networks,
  Schur decomposition, nonnormality}
  \vskip 0.3in
]

\printAffiliationsAndNotice{}

\begin{abstract}
  Recurrent networks can contain substantial functional redundancy in
  weight space: changing a recurrent matrix may leave the input-output
  rollout nearly unchanged on a task distribution, while similar-scale
  changes can destroy the same behavior. We study this redundancy in
  one-layer tanh RNNs using ordered real Schur coordinates. The Schur
  form separates spectral blocks from directed nonnormal couplings,
  giving a diagnostic basis for structured ablations that keep the input
  and readout maps fixed. In a fixed-length copy task, selected
  nonnormal Schur couplings can be removed with little loss in some
  trained solutions, whereas other couplings are necessary for accurate
  autonomous replay. Across flip-flop, sine generation, and
  context-dependent integration, the loss-preserving ablation profile
  varies across tasks and trained solutions. These results identify
  candidate approximate functional invariances, not universal symmetries
  of recurrent weight space. Schur-coordinate ablations provide a
  practical diagnostic for which structured perturbations preserve a
  trained recurrent solution and which ones disrupt its computation.
\end{abstract}

\section{Introduction}

Exact weight-space symmetries have become a practical tool for
comparing neural networks and for learning directly in parameter
space~\cite{entezari2022role,ainsworth2023git,navon2023equivariant,navon2023deepalign}.
Those symmetries identify transformations that preserve the realized
function exactly, and recent work builds such structure directly into
models that operate on trained networks as
inputs~\cite{zhou2023permutation,kofinas2024graph}. Recurrent
networks can also admit large structured changes to the recurrent
matrix that preserve task behavior only approximately and only on the
task distribution. These directions fall outside exact group-theoretic
symmetries, while still shaping the functional geometry of weight
space.

Ordered Schur coordinates reveal candidate approximate functional
invariances under structured perturbation. Because the resulting
ablation profiles vary by task and by trained solution, they should not
be read as evidence that nonnormal components can usually be ignored.
They identify which Schur-coordinate couplings a particular recurrent
solution can lose while preserving its original input-output rollout,
and which couplings carry task-specific function.

Because tanh RNNs do not admit arbitrary orthogonal changes of basis as
exact symmetries, raw recurrent coordinates make nonnormal structure
hard to compare across runs. The real Schur decomposition represents
every real recurrent matrix by an orthogonal basis, diagonal or
quasi-diagonal spectral blocks, and strictly upper-triangular nonnormal
interactions.
Such interactions are known to shape transient recurrent
computations~\cite{murphy2009balanced,hennequin2012nonnormal,bondanelli2020coding,pattadkal2024primate},
and ordered Schur coordinates make them comparable and ablatable.

Schur-coordinate ablations preserve the rollout function for some
blocks and not for others. In the copy task, selected ablations produce
nearly identical autonomous replay accuracy, while directed
cross-sector ablations move the model to lower-accuracy behavior. The
neuroscience-style tasks provide a scope test for the same
interventions. The copy task supplies an explicit temporal symmetry; the
flip-flop, sine-generation, and context-dependent integration tasks ask
whether the same diagnostic basis also localizes fragile directions in
other recurrent computations~\cite{sussillo2013opening,mante2013context,maheswaranathan2019universality,schuessler2024aligned}.
Task-dependent ablation profiles tie approximate invariance to the
rollout distribution rather than to a task-independent property of Schur
blocks.

\section{Ordered Schur Coordinates}

A one-layer tanh RNN maps input
\(x_t \in \mathbb{R}^{N_x}\), hidden state
\(h_t \in \mathbb{R}^{N_h}\), output
\(\hat{y}_t \in \mathbb{R}^{N_y}\),
\begin{align}
  h_t &= \tanh(W_{xh} x_t + W_{hh} h_{t-1}), \qquad h_0 = 0, \\
  \hat{y}_t &= W_{hy} h_t,
\end{align}
with \(W_{xh} \in \mathbb{R}^{N_h \times N_x}\),
\(W_{hh} \in \mathbb{R}^{N_h \times N_h}\), and
\(W_{hy} \in \mathbb{R}^{N_y \times N_h}\). All reported experiments
set the recurrent and readout biases to zero, \(b_h=b_y=0\).

For a trained recurrent matrix, write \(W=W_{hh}\). Its real Schur
decomposition is
\begin{equation}
  W = Q T Q^\top,
  \label{eq:schur}
\end{equation}
where \(Q\) is orthogonal and \(T\) is real quasi-upper-triangular
\cite{trefethen2005spectra}. We decompose
\begin{equation}
  T = B + N,
  \label{eq:schur_bn}
\end{equation}
where \(B\) contains the block-diagonal \(1\times1\) and \(2\times2\)
real Schur eigenvalue blocks, and \(N\) contains the strictly
block-upper-triangular nonnormal couplings between those blocks.

The Schur blocks are ordered by nonincreasing eigenvalue modulus. A
relative threshold \(\alpha\) separates leading spectral blocks from
their complement:
\[
  R=\{i:|\lambda_i|\ge \alpha\rho(W)\},\qquad
  C=\{1,\ldots,N_h\}\setminus R .
\]
Here \(\lambda_i\) is the eigenvalue associated with the \(i\)th Schur
block and \(\rho(W)=\max_j|\lambda_j|\) is the spectral radius of
\(W\). \(R\) indexes the leading rotation-like subspace used as the
reference sector, while \(C\) indexes the remaining Schur blocks whose
couplings to \(R\) and to each other are tested by ablation. In this
ordered partition,
\begin{equation}
  B =
  \begin{pmatrix}
    B_R & 0 \\
    0 & B_C
  \end{pmatrix},
  \qquad
  N =
  \begin{pmatrix}
    \RR & \CR \\
    0 & \CC
  \end{pmatrix}.
  \label{eq:schur_partition}
\end{equation}
\(\RR\), \(\CR\), and \(\CC\) are blocks of the nonnormal coupling
matrix \(N\), not separate eigenvalue blocks. The cross block \(\CR\)
is the upper-right coupling from the complement sector into the leading
sector in the ordered Schur coordinates.

For a set \(S\) of Schur-coupling blocks, the intervention zeros the
corresponding entries of \(N\), reconstructs
\begin{equation}
  \widetilde{W}_{hh}(S) = Q \widetilde{T}(S) Q^\top,
  \label{eq:schur_intervention}
\end{equation}
and reevaluates the original network without changing input or readout
weights. Let \(f_W\) denote the rollout function of the trained network
on a task distribution \(\mathcal D\). This fixed-encoder/fixed-decoder
intervention tests whether the original input-output map is preserved
in the original readout coordinates. Refitting a linear or ridge
decoder after the ablation would answer a different question: whether
the perturbed latent dynamics still contain task information up to a
new readout.

For a rollout discrepancy \(d_{\mathcal D}\) and tolerance
\(\epsilon\), an intervention \(S\) is an \(\epsilon\)-stabilizer on
\(\mathcal D\) when
\(d_{\mathcal D}(f_W,f_{\widetilde{W}_{hh}(S)})\le \epsilon\). A
Schur-coupling block is a candidate approximate functional invariance
when zeroing it gives small discrepancy while removing non-negligible
Schur mass. If performance changes sharply, the block lies in a
fragile functional direction for that trained solution.

For neuroscience-style tasks, held-out error is measured by
\begin{equation}
  \fvu = \frac{\mathbb{E}\|\hat{y}-y\|^2}{\mathbb{E}\|y-\bar{y}\|^2}.
\end{equation}
The expectation is over held-out rollouts, \(y\) is the target
trajectory, \(\hat{y}\) is the model output, and \(\bar{y}\) is the
empirical mean target over the evaluation set.
For those tasks two summaries are reported:
\begin{align}
  \Delta \fvu &= \fvu(\widetilde{W}_{hh}) - \fvu(W_{hh}), \\
  S_{\Delta T} &= \frac{\Delta \fvu}{\|\Delta T\|_F / \|T\|_F}.
\end{align}
\(\Delta T=T-\widetilde T(S)\), and \(\|\cdot\|_F\) denotes the
Frobenius norm. \(\Delta \fvu\) captures the effect at the trained
scale, whereas \(S_{\Delta T}\) measures effect per unit removed Schur
mass. The perturbations are evaluated after training; no input or
readout weights are refit.

We use \(\alpha=0.9\) throughout the main experiments. This value was
chosen a priori as a simple relative spectral-radius cutoff for
grouping high-modulus Schur blocks into \(R\), rather than tuned for an
ablation outcome. The threshold controls only the \(R/C\) partition
used to assign nonnormal couplings to \(\RR\), \(\CR\), and \(\CC\).
A nearby-threshold check on the copy-task controllers preserves the
same qualitative profile (\cref{tab:alpha_sensitivity}).

\begin{table}[t]
  \centering
  \caption{Sensitivity to the Schur split threshold. Values are mean
    autonomous replay accuracy over 128 lags. The main experiments use
    \(\alpha=0.9\).}
  \label{tab:alpha_sensitivity}
  \begin{tabular}{@{}llcccc@{}}
    \toprule
    \(\alpha\) & model & \(n_R\) & full & \(-\CC\) & \(-\CR\) \\
    \midrule
    0.85 & dense orth. & 64 & 1.000 & 1.000 & 0.634 \\
    0.90 & dense orth. & 64 & 1.000 & 1.000 & 0.634 \\
    0.95 & dense orth. & 64 & 1.000 & 1.000 & 0.634 \\
    0.85 & Cayley & 68 & 1.000 & 1.000 & 1.000 \\
    0.90 & Cayley & 68 & 1.000 & 1.000 & 1.000 \\
    0.95 & Cayley & 68 & 1.000 & 1.000 & 1.000 \\
    \bottomrule
  \end{tabular}
\end{table}

\paragraph{Coordinate choice.}
The Schur basis remains orthogonal even for strongly nonnormal
matrices~\cite{trefethen2005spectra}. Direct eigencoordinates are
often ill-conditioned when transient amplification is large, making
cross-run comparison unstable and turning component ablations into
basis-sensitive operations. By separating spectral blocks from
nonnormal couplings and ordering them by eigenvalue modulus, the real
Schur form turns those couplings into structured perturbation
directions. Compared with eigencoordinates, Schur coordinates provide a
reproducible diagnostic basis for perturbing and interpreting recurrent
dynamics.

\section{Approximate Stabilizers in the Copy Task}

The copy task is a fixed-delay variant of the copying-memory benchmark
for long-range recurrent memory~\cite{hochreiter1997long,arjovsky2016unitary},
and related fixed-length copy tasks have been used to study
traveling-wave recurrent models~\cite{keller2024traveling}. It
presents a sequence of \(s=8\) symbols in
\(\{-1,+1\}^d\), with \(d=8\), then sets inputs to zero while the network
autonomously reproduces the stored sequence. Replay accuracy is
measured over the first 128 generated symbols after the input sequence.
The copy task experiments train one-layer
tanh RNNs at \(N_h \in \{56,64,72\}\) under four recurrent constructions.
Let \(m=N_h^{-1/2}\). The three dense constructions optimize an
unconstrained matrix \(W_{hh}\in\mathbb{R}^{N_h\times N_h}\) and
differ only in \(W_{hh}^{(0)}\):
\[
  \begin{aligned}
    \text{dense default:}\quad
    &W_{hh,ij}^{(0)} \sim \mathrm{Unif}[-m,m],\\
    \text{dense orthogonal:}\quad
    &W_{hh}^{(0)} = Q,\qquad Q^\top Q=I,\\
    \text{dense normal:}\quad
    &W_{hh}^{(0)} = QD_{\mathrm{norm}}Q^\top ,
  \end{aligned}
\]
where
\[
  \begin{aligned}
    D_{\mathrm{norm}}&=\mathrm{blockdiag}(B_1,\ldots,B_{N_h/2}),\\
    B_i&=
    \begin{pmatrix}
      a_i & -b_i\\
      b_i & a_i
    \end{pmatrix},\\
    a_i,b_i&\sim\mathcal{N}(0,1/6).
  \end{aligned}
\]
For the Cayley construction, every optimization iterate satisfies
\(W_{hh}^{(k)}=O(A^{(k)})D^{(k)}O(A^{(k)})^\top\), where
\((A^{(k)})^\top=-A^{(k)}\) and
\[
  O(A) = (I-A)(I+A)^{-1}.
\]
At initialization,
\[
  \begin{aligned}
    U_{ij}&\sim\mathrm{Unif}[-m,m],\\
    A^{(0)}&=(U-U^\top)/2,\\
    \widetilde{W}_{ij}&\sim\mathrm{Unif}[-m,m],\\
    D^{(0)}&=\mathrm{realblock}\!\left(\mathrm{eig}(\widetilde{W})\right),
  \end{aligned}
\]
with \(\mathrm{realblock}(\cdot)\) converting conjugate eigenvalue
pairs into \(2\times2\) real blocks of the form above.
For \(\mathcal Z=\{\RR,\CR,\CC\}\) and \(S\subseteq\mathcal Z\), the
intervention is
\[
  \begin{aligned}
    \widetilde{W}_{hh}(S) &= QZ_S(T)Q^\top,\\
    \bigl(Z_S(T)\bigr)_B &=
    \begin{cases}
      0, & B\in S,\\
      T_B, & B\notin S,
    \end{cases}
    \qquad B\in\mathcal Z.
  \end{aligned}
\]
Entries outside \(\{\RR,\CR,\CC\}\) are unchanged. For
\(\mathcal D_{\mathrm{rc}}\) and \(\mathcal L=\{1,\ldots,128\}\),
\[
  \hat{y}_{\ell j}^S(x) := \hat{y}_{\ell j}(x;\widetilde{W}_{hh}(S)),
\]
\[
  \begin{aligned}
  \mathrm{Acc}_{\mathrm{rc}}
  &=
  \frac{1}{|\mathcal D_{\mathrm{rc}}|\,|\mathcal L|\,d}
  \sum_{\substack{(x,y)\in\mathcal D_{\mathrm{rc}}\\
                  \ell\in\mathcal L,\; j\in[d]}}
  \mathbf 1\{\operatorname{sgn}(\hat{y}_{\ell j}^S(x))=y_{\ell j}\}.
  \end{aligned}
\]

\begin{figure}[t]
  \centering
  \includegraphics[width=0.98\columnwidth]{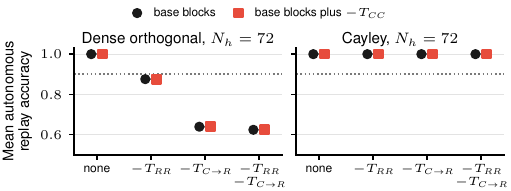}
  \caption{Candidate approximate functional invariances in the copy task. Points connected by
    gray line segments differ only by additionally zeroing \(\CC\).
    In the dense orthogonal model, \(\CC\) removal leaves the
    autonomous replay function nearly unchanged conditional on the
    other removed blocks, while \(\RR\) and \(\CR\) move the network
    between lower-accuracy functional classes. The Cayley-transform
    representative has negligible complement blocks and changes little
    under the shown ablations.}
  \label{fig:repeatcopy_equivalence}
\end{figure}

In the dense orthogonal \(N_h=72\) model, removing \(\CC\) alone
leaves mean replay accuracy at \(1.00\), matching the full model
(\cref{fig:repeatcopy_equivalence}). The same near-equivalence holds
after other Schur blocks have already been removed: \(-\RR\) and
\(-\RR,-\CC\) give \(0.876\) and \(0.875\); \(-\CR\) and
\(-\CR,-\CC\) both give \(0.639\); \(-\RR,-\CR\) and zeroing all
three blocks both give \(0.624\). Selected structured changes to
nonnormal Schur couplings can therefore preserve the task behavior
once the other ablated blocks are fixed.

\(\CC\) is close to a stabilizer for this solved copy task
controller conditional on the other removed blocks. Removing \(\CR\)
moves the dense model to a different functional class, and removing
\(\RR\) produces a distinct intermediate class. The Cayley representative has
negligible complement blocks at this width, so the same ablations
leave replay accuracy unchanged.

These pairs define task-restricted approximate equivalence classes in
which multiple recurrent matrices with different nonnormal
coordinates realize nearly identical rollout functions on the
copy task distribution. The copy task panels evaluate two
representative solved \(N_h=72\) controllers, using Schur-coordinate ablations as mechanistic
interventions on trained controllers.

\emph{Takeaway:} in the dense orthogonal copy solution, \(\CC\) is
nearly loss-preserving while \(\CR\) is not; in the Cayley-transform
solution, the tested nonnormal couplings are nearly absent and the same
ablations have little effect.

\section{Task Dependence Beyond the Copy Task}

The cross-task suite tests whether the Schur-coordinate interventions
remain informative beyond the explicit temporal symmetry of the copy task.
The three tasks require discrete memory, oscillatory generation, and
context-dependent accumulation, so they probe distinct recurrent
computations in the same one-layer architecture. The experiments use
one-layer tanh RNNs with
\(N_h=64\), \(W_{hh}^{(0)}=Q\), \(Q^\top Q=I\), Adam with learning
rate \(10^{-3}\), batch size 64, 30
epochs, and 128 batches per epoch. Three seeds are trained for each
of 3-bit flip-flop with length 25, frequency-cued sine generation with
length 50, and context-dependent integration with four inputs and
length 48. Full models have held-out mean \(\fvu = 0.0048\) for
flip-flop, \(0.0036\) for sine generation, and \(0.0104\) for
context-dependent integration.

\begin{figure}[t]
  \centering
  \includegraphics[width=0.98\columnwidth]{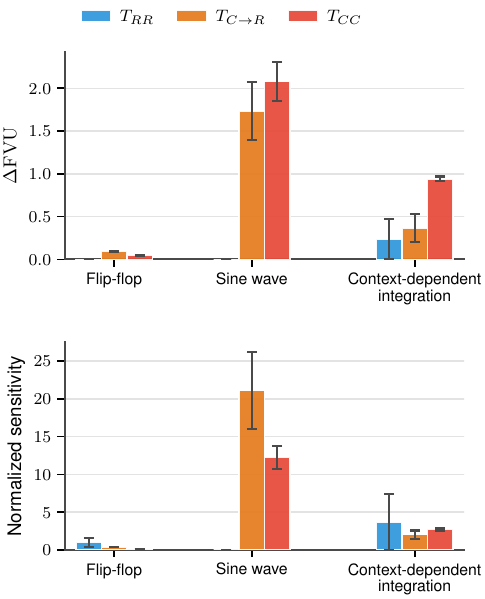}
  \caption{Single-block Schur ablations across neuroscience-style
    tasks. Top: raw degradation \(\Delta \fvu\). Bottom: normalized
    sensitivity \(S_{\Delta T}\). The loss-preserving ablation profile
    depends on the computation: raw degradation is largest for \(\CR\)
    in flip-flop and for complement-linked blocks in sine generation and
  context-dependent integration.}
  \label{fig:sensitivity}
\end{figure}

Values are mean \(\pm\) SEM over seeds (\cref{fig:sensitivity}).
For flip-flop, zeroing \(\CR\) increases held-out error by
\(9.45 \times 10^{-2} \pm 9.35 \times 10^{-3}\), while zeroing
\(\CC\) increases error by \(4.96 \times 10^{-2} \pm 5.39 \times
10^{-3}\). The ring-internal block \(\RR\) has almost no raw effect.

For sine generation, zeroing \(\CC\) raises held-out error by \(2.08
\pm 0.23\), and zeroing \(\CR\) raises it by \(1.73 \pm 0.34\). The
normalized sensitivity is largest for \(\CR\), \(21.1 \pm 5.1\), with
\(\CC\) still substantial at \(12.3 \pm 1.5\). Removing \(\RR\) has little
effect at this width. In context-dependent integration,
zeroing \(\CC\) raises held-out error by \(0.94 \pm 0.03\), while
zeroing \(\CR\) raises it by \(0.37 \pm 0.16\). The raw effect is
dominated by \(\CC\), consistent with a slow accumulated variable
supported by within-complement recurrence.

Across tasks, selected Schur couplings can be removed with little loss
when they avoid task-relevant directions, as in the copy task \(\CC\)
pairs. The same coordinates localize fragile directions when a block is
required, as in sine generation and context-dependent integration.

\emph{Takeaway:} the loss-preserving ablation profile varies across
tasks and trained solutions; no single Schur coupling is uniformly safe
to remove.

\paragraph{Metric interpretation.}
Raw degradation measures loss at the trained operating point, whereas
\(S_{\Delta T}\) measures loss per unit removed Schur mass. We treat
\(\Delta\fvu\) as the primary behavioral effect and use
\(S_{\Delta T}\) to identify small sectors with disproportionate
impact.

\section{Discussion and Limitations}

\paragraph{Interpretation.}
Exact symmetries characterize functional equivalence classes in weight
space~\cite{entezari2022role,ainsworth2023git,navon2023equivariant}.
Ordered Schur coordinates play a complementary role by fixing an
orthogonal coordinate system in which recurrent matrices can be compared
and nonnormal sectors can be causally ablated. The resulting equivalences
are task-restricted and approximate, because they are defined by
rollout behavior on \(\mathcal D\) rather than by a global
parameter-space group action. For recurrent networks, raw parameter
distance can miss both large structured changes that preserve the
task function and small directed changes that alter it.

Because the tasks studied here are low-dimensional, the trained
networks may use only a low-dimensional hidden-state subspace. A Schur
ablation can then preserve performance because it avoids the activity
directions aligned with the readout or the dominant hidden-state
principal components, rather than because the removed coupling has no
computational role.
The experiments do not separate this subspace explanation from the
Schur-coordinate account. Separating the two would require measuring how
the ablated Schur directions project onto hidden-state PCs,
readout-aligned subspaces, and task-conditioned activity manifolds.

\paragraph{Scope.}
The experiments use vanilla one-layer tanh RNNs, simple
low-dimensional tasks, a narrow width range, and a small number of
trained solutions. They do not test LSTMs, GRUs, gated architectures,
large sequence models, or high-dimensional real-world sequence tasks, so
the evidence supports Schur-coordinate ablation as a diagnostic for
trained recurrent controllers rather than a universal statement about
nonnormal structure.

\bibliographystyle{icml2026/icml2026}
\bibliography{references}

\end{document}